\documentclass[sigconf]{acmart}

\AtBeginDocument{%
  \providecommand\BibTeX{{%
    \normalfont B\kern-0.5em{\scshape i\kern-0.25em b}\kern-0.8em\TeX}}}

\setcopyright{acmcopyright}
\copyrightyear{2018}
\acmYear{2018}
\acmDOI{XXXXXXX.XXXXXXX}

\acmConference[BIOKDD '22]{21th International Workshop on Data Mining in Bioinformatics}{Aug 15,
 2022}{Washington DC, USA}
\acmPrice{15.00}
\acmISBN{978-1-4503-XXXX-X/18/06}

\begin{document}

\title{DDI Prediction via Heterogeneous Graph Attention Networks}


\author{Farhan Tanvir}
\affiliation{%
  \institution{Oklahoma State University}
  \city{Stillwater}
  \state{Oklahoma}
  \country{USA}}
\email{farhan.tanvir@okstate.edu}

\author{Khaled Mohammed Saifuddin}
\affiliation{%
  \institution{Oklahoma State University}
  \city{Stillwater}
  \state{Oklahoma}
  \country{USA}
}
\email{khaled_mohammed.saifuddin@okstate.edu}

\author{Esra Akbas}
\affiliation{%
 \institution{Oklahoma State University}
 \streetaddress{Rono-Hills}
 \city{Stillwater}
 \state{Oklahoma}
 \country{India}}
\email{eakbas@okstate.edu}
\renewcommand{\shortauthors}{Tanvir, et al.}

\begin{abstract}

Polypharmacy, defined as the use of multiple drugs together, is a standard treatment method, especially for severe and chronic diseases. However, using multiple drugs together may cause interactions between drugs. Drug-drug interaction (DDI) is the activity that occurs when the impact of one drug changes when combined with another. DDIs may obstruct, increase, or decrease the intended effect of either drug or, in the worst-case scenario, create adverse side effects. While it is critical to detect DDIs on time, it is time-consuming and expensive to identify them in clinical trials due to their short duration
and many possible drug pairs to be considered for testing. As a result, computational methods are needed for predicting DDIs. In this paper, we present a novel heterogeneous graph attention model, \texttt{HAN-DDI} to predict drug-drug interactions. We create a heterogeneous network of drugs with different biological entities. Then, we develop a heterogeneous graph attention network to learn DDIs using relations of drugs with other entities. It consists of an attention-based heterogeneous graph node encoder for obtaining drug node representations and a decoder for predicting drug-drug interactions. Further, we utilize comprehensive experiments to evaluate of our model and to compare it with state-of-the-art models. Experimental results show that our proposed method, \texttt{HAN-DDI}, outperforms the baselines significantly and accurately predicts DDIs, even for new drugs.
\end{abstract}




\maketitle

\section{Introduction}
Taking multiple drugs together, called a polypharmacy, has become a highly successful method of treating diseases, especially for severe and chronic diseases. \cite{Han2017SynergisticDC,  Liebler2005ElucidatingMO}. On the other hand, polypharmacy may cause substantial adverse drug reactions (ADR) owing to interaction between drugs~\cite{Han2022}. Drug-drug interactions (DDIs) may alter the activity of the drugs with obstructing, increasing, or decreasing the intended
effect of drug, and in the worst-case scenario, may create adverse
side effects~\cite{Grando2012OntologicalAF}. To mitigate unintended pharmacological side-effects' impact, it is critical to predict DDIs early and effectively. However,  it is difficult to identify all potential DDIs, especially for new drugs  with clinical trials due to the performing on a minimal number of patients in 
a short duration and many possible drug pairs to be considered for testing. Thus, many computational models are developed to detect DDIs automatically. 

Most DDI prediction approaches integrate numerous data sources for drug attributes such as similarity features (\cite{Ma2018DrugSI, Ryu2018DeepLI}), adverse or side effects (\cite{ZitnikAL18, Jin2017MultitaskDP}), and multi-task learning (\cite{Chu2019MLRDAAM}). These techniques rely on the hypothesis that similar drugs have a similar type of interaction. Meanwhile, other computational techniques utilize popular embedding methods (\cite{Quan2018ASF, Le2018ACVtreeAN, Quan2019AnEF}), which learn drug representation and perform DDI prediction. Although the abovementioned approaches show excellent performance, they ignore that DDI is represented as an independent data sample, and associated relationships are not taken into account.

Due to the widespread use of knowledge graphs (KG), there has been an increase in study on relation inference and recommendation. Recent research has employed KG to predict DDI~\cite{Karim2019DrugDrugIP, elebi2019EvaluationOK}. To extract drug features using different embedding techniques, they both apply KG to machine learning models. They do learn node latent embedding directly. However, these techniques are limited in their ability to retrieve the detailed neighborhood. Recently, graph neural networks have shown to be very effective for DDI prediction \cite{ZitnikAL18, Lin2020KGNNKG} through effectively capturing local neighborhood and graph structure. These methods take interaction of drugs and other biomedical entities into account to form an enhanced node representation. However, they do not consider significance of different type of interactions and treat each kind of interactions equally.

Recently, many studies have been using relationships between drugs and other biological entities. A question might arise- how do we model these complete, enriched data? To address this, networks, which are graphs encapsulating the intricate structure of interactions between related entities can be used. Networks are used in various areas, including social networks~\cite{akbas2017truss, akbas2020proximity}, citation networks~\cite{tanner2019paper}, and biological networks~\cite{Tanvir2021PDDI, Bumgardner2021DrugDrugIP}. To represent various entities and their disparate interactions, heterogeneous Information Network (HIN)\cite{Shi2017ASO} is defined. It is used in a variety of applications including  DDI prediction ~\cite{Tanvir2021PDDI,  Karim2019DrugDrugIP} and detecting opioid addicts from social media ~\cite{metapathopu}. 

In this paper, we present a novel heterogeneous graph attention networks model for the DDI prediction problem. First, we extract extensive drug-centric information, comprising data on drugs, different biomedical entities, and their diverse and multi-relational interactions, from DrugBank.
Then, we model these data using a Heterogeneous Information Network (HIN). Then, we construct different meta-paths connecting different drugs in our HIN to define similarities between drugs. Meta-paths provide a way to denote how connected or relevant two drugs are, based on their interaction with other entities like protein and disease. 

Furthermore, we develop heterogeneous graph attention neural network (HAN) model (\cite{Wang2019HeterogeneousGA}) for DDI prediction. \texttt{HAN-DDI} has an encoder-decoder architecture. We construct a hierarchical attention mechanism in the encoder that includes node-level and meta-path-level attention to learn node representations from various meta-paths. This model propagates information from local neighbors using a meta-path. For each meta-path, we integrate node-level attention to learn representations for nodes. We also use meta-path-level attention to learn the importance of distinct meta-paths to efficiently aggregate node representations from different meta-paths. Afterward, the pair-wise representations of drugs are passed through the decoder function to predict a binary score for each drug pair that indicates whether two drugs interact. To the best of our knowledge, this is the first method to solve DDI prediction using a hierarchical attention mechanism, which successfully learns node embedding from numerous meta-paths.

As the initial features of drugs to give as the input to our model in addition to relational information, we extract comprehensive features for drugs. Drugs are massive chemical compounds. While numerous chemical substructures exist in a drug, only a few chemical substructures cause chemical reactions among drugs. Therefore, we use the ESPF algorithm (\cite{espf}) for feature extraction that retrieves frequent substructures from a drug’s chemical structure.

We conduct extensive experiments to compare our model with the state-of-the-art models. \texttt{HAN-DDI} comprehensively outperforms other state-of-the-art models by up to 19\%. We also depict supporting evidence in the biomedical literature for our novel predictions, indicating that \texttt{HAN-DDI} excels at predictions that are likely to be true positive.

Our primary contributions are as follows:
\begin{itemize}
\item \textbf{Drug-centric Interaction Integration on HIN}:  We construct a Heterogeneous Information Network (HIN) with taking into account drugs and other biomedical entities like proteins, side effects, and chemical structures.

\item \textbf{Constructing Novel Meta-paths}: We utilize different meta-paths considering drugs’ interaction with other biomedical entities.  

\item \textbf{Heterogeneous Graph Encoder-Decoder}: We introduce a novel graph encoder-decoder framework for predicting drug-drug interactions based on a heterogeneous graph attention network. 

\item \textbf{Extensive Experiments}: We perform extensive experiment  to compare our model with the state-of-the-art models.  We also design experiments for new drugs, and experimental results demonstrate that our model can predict drugs with no known interactions.

\end{itemize}

The structure of this paper is outlined as follows. In this section, we explain the DDI problem and its impact. In addition, we explore related works in Section \ref{works}. Section \ref{sec:method} describes how data
from various sources are integrated, creating a heterogeneous graph. Moreover, we elaborate on the \texttt{HAN-DDI} method, consisting of an encoder-decoder framework. Furthermore, we describe our experiments and illustrate our results in Section \ref{experiment}. Finally, we conclude in Section \ref{conclusion}.

\section{Related Works}
\label{works}
Previous DDI prediction research may be divided into similarity-based and deep learning-based techniques.

\subsection{Similarity-based Methods}
To predict ADRs, pharmacological, topological, or meta-path similarity based on statistical learning is traditionally calculated ~\cite{elebi2015PredictionOD, Kastrin2018PredictingPD}. Similarity-based approaches have been shown to be successful in predicting drug-drug interactions (DDIs). These strategies are based on the notion that similar drugs will interact with one another. Various research publications (\cite{Gottlieb2012INDIAC, Abdelaziz2017LargescaleSA}) used several numbers and types of similarity metrics to predict DDIs. Another research paper worth mentioning is \cite{Tanvir2021PDDI}, which used many data sources to construct a heterogeneous graph. Furthermore, they acknowledged that their dataset was imbalanced and skewed and devised many experiments to solve these drawbacks. Meanwhile, \cite{Bumgardner2021DrugDrugIP} used a hypergraph to depict chemical structure-based similarity between drugs. In this work, multiple drugs can share a hyper-edge if drugs share a similar chemical substructure. However, the majority of these methods take into account fewer datasets and drug-centric interactions. 

\subsection{Deep Learning-based Methods}
Recently, a growing number of research-based neural networks, especially graph neural networks, have addressed different problems, including DDI prediction~\cite{feng2022prediction} and drug abuse detection \cite{Saifuddin2021DrugAD}. Graph neural network-based approaches construct knowledge graphs based on drug-centric interactions. Afterward, they employ a graph neural network to extract relations among drugs. Decagon \cite{ZitnikAL18} constructed a knowledge graph based on protein-protein interactions, drug-drug interactions, and drug-protein interactions. Afterward, they developed a graph convolutional network consisting of encoding, decoding, and model training phases for DDI prediction. KGNN \cite{Lin2020KGNNKG} used GNN to learn drugs and their embedding using a knowledge graph and DDI. CASTER \cite{Huang2020CASTERPD} recently developed a dictionary learning framework for predicting DDIs given drug chemical structures via SMILES string. HyGNN \cite{Saifuddin2022HyGNN}, a hypergraph neural network, is composed of a novel attention-based hypergraph edge encoder for representing drugs as hyperedges and a decoder for predicting drug interactions. Although these techniques have demonstrated excellent performance, one fact they ignore is that they treat DDI as an independent data sample and do not take into account their relationships in the knowledge graph. The main distinction between our study and the literature is that we attempt to extract interactions between drugs and other biomedical entities utilizing various meta-paths. Furthermore, we use the attention mechanism to obtain the weight of drug nodes and the meta-paths.

\section {The Proposed model }
\label{sec:method}

\begin{figure*}[t]
    \centering
    \includegraphics[width=\textwidth]{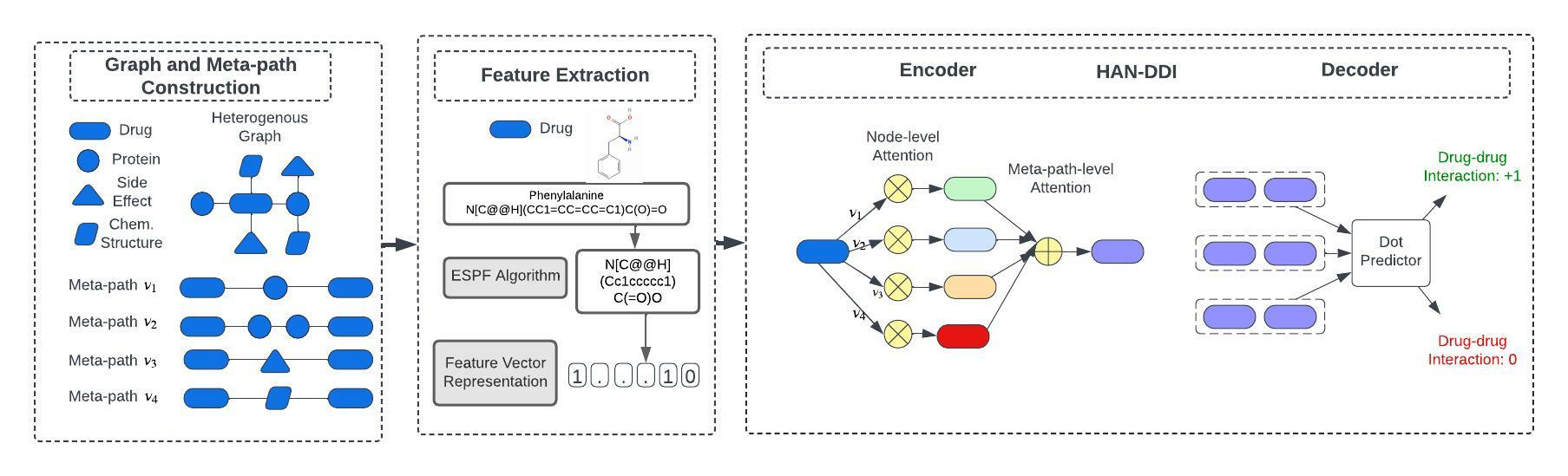}
   \caption{System Architecture of HAN-DDI}
    \label{fig:sys_arch}
\end{figure*}

Given two drugs $d_i$ and $d_j$, our goal is to predict whether these two drugs interact or not. In this paper, we model our data as a Heterogeneous Information Network (HIN) and create different meta-paths connecting different drugs in our HIN. Then, we develop a multi-layer heterogeneous graph attention network \texttt{HAN-DDI}, a model including an encoder to produce features of drugs.
The encoder has a hierarchical attention structure: node-level attention and meta-path-level attention. Figure ~\ref{fig:sys_arch} depicts the entire \texttt{HAN-DDI} structure. To begin, we present a node-level attention method for learning the weight of meta-path based neighbors and aggregating them to obtain the meta-path-specific node embedding. Following that, \texttt{HAN-DDI} may distinguish between meta-paths using meta-path-level attention and obtain the ideal weighted combination of meta-path-specific node embedding. Finally, \texttt{HAN-DDI} incorporates a decoder that learns and predicts the interaction between drugs.

Our proposed model consists of the following steps:
\begin{enumerate}
\item Heterogeneous Graph Construction
\item Meta-path construction
\item Feature Extraction
\item DDI prediction with Heterogeneous graph attention network 
\begin{itemize}
    \item Encoder: Attention-based Drug representation learning
    \item Decoder: DDI learning

\end{itemize}
\end{enumerate}

\subsection{Heterogeneous Graph Construction}

To predict DDIs, we follow the underlying hypothesis that similar drugs interact with each other. To properly define drug similarities, we consider interactions between drugs and other bio-molecular entities. We aim to compute similarity between drugs based on their interaction with other biomedical entities.

We consider drugs, proteins, side-effect as the entities and use DrugBank data to build relationships between these entities. We also extract the  substructures from chemical structure of drugs and define it as another entity and relation in the network. Detailes are given below.

\begin{itemize}
    \item \textbf{I1:} $T$ matrix represents the drug-target protein interaction where each element $t_{i,j}$ states whether drug \textit{i} targets protein \textit{j}.
    
    \item \textbf{I2:} $SE$ matrix represents drug-side effect relation where each element $se_{i,j}$ describes whether drug \textit{i} causes side effect \textit{j}.
    
    \item \textbf{I3:} $H$ matrix outlines drug-chemical substructure relation where each element $h_{i,j}$ refers to whether drug \textit{i} have chemical substructure \textit{j}.
    
    \item \textbf{I4:} To represent protein-protein interaction (PPI) interaction, we build the protein-interacts-protein matrix $P$, where each entry $p_{i,j}$ indicates whether or not protein pair \textit{i} and \textit{j} interacts or not.

\end{itemize}

Table ~\ref{tab1} summarizes the relations mentioned above and their elements in the relation matrices.
To integrate all these interactions between these diverse entities, we create heterogeneous information networks.

\begin{definition}[Heterogeneous information network] \cite{Shi2017ASO} A heterogeneous information network (HIN) is defined as a graph $\mathrm{G}$ = $(\mathrm{V}$,$\mathrm{E})$ with an entity type mapping $\mathrm{\phi}$: V $\mathrm{\rightarrow}$ $\mathrm{A}$ and a relation type mapping  $\mathrm{\psi}$ : $\mathrm{\epsilon}$ $\mathrm{\rightarrow}$ $\mathrm{R}$, where $\mathrm{V}$ denotes the entity set and $\mathrm{E}$ is the relation set, $\mathrm{A}$ denotes the entity type set and $\mathrm{R}$ is the relation type set and the number of entity types $\mathrm{|A|>1}$ or the number of relation type $\mathrm{|R|>1}$.
\end{definition}


HIN offers the network structure of data associations and a high-level abstraction of the categorical association. As stated previously, we have four entity kinds (drug, protein, chemical structure, and side effects) and four types of relationships among them for DDI prediction.

 \subsection{Meta-path construction}

After constructing the heterogeneous network, we create \textit{meta-paths} \cite{Sun2011PathSimMP} to extract relations among drugs through other entities. For a heterogeneous graph, meta-paths are used to measure the relationships and similarities between entities. Furthermore, meta-paths are represented by a commuting matrix. Meta-path and commuting matrix are defined below.

\begin{table*}[h]
    \centering
      \caption{Description of Each Matrix And Its Element}
      \begin{tabular}{ccc }
      \toprule
       \textbf{Matrix} & \textbf{Element} & \textbf{Description}  \\
       \midrule
        T  & $t_{i,j}$ & if drug $i$ targets protein $j$, then $t_{i,j}$ = 1; otherwise, $t_{i,j}$=0
        \\         
        C & $c_{i,j}$ & if drug $i$ causes side-effect $j$, then $c_{i,j}$ = 1;otherwise, $c_{i,j}$=0
        \\       
        H & $h_{i,j}$ & if drug $i$ possesses chemical sub-structure $j$, then $h_{i,j}$ = 1;otherwise, $h_{i,j}$=0
        \\
        P & $p_{i,j}$ & if protein pair $i$ and $j$ interacts, then $p_{i,j}$ = 1;otherwise, $p_{i,j}$=0
        \\
         
        \bottomrule
        \end{tabular}
        \label{tab1}
     \end{table*}

\begin{definition}[Meta-path] \cite{Sun2011PathSimMP} A meta-path $P$ is a path on the network schema diagram $T_G$ = $(A, R)$, and is represented in the shape of $A_1$ $\xrightarrow{R_1}$ $A_2$ $\xrightarrow{R_2} \cdots \xrightarrow{R_l}$ $A_{L+1}$, describing a composite relationship $R = R1 \circ R2 \circ \cdots \circ R$ between entities $A_1$ and $A_{L+1}$, where $\circ$ denotes composition operator association, and length of $\mathrm{P}$ is $\mathrm{L}$. 
\end{definition}

\begin{definition}[Commuting matrix] \cite{Sun2011PathSimMP} Given a network $G$, a commuting matrix $M_P$ for a meta-path $P$ = ($A_1A_2\cdots A_{L+1}$) is defined as $M_P$ = $(G_{A_1A_2} G_{A_2A_3} \cdots$ $G_{A_lA_{L+1}})$, where $G_{A_iA_j}$ is the adjacency matrix between types $A_i$ and $A_j$. $M_P (i, j)$ represents the number of path instances between entity $x_i$ $\in$ $A_1$ and entity $y_i$ $\in$ $A_{L+1}$ under meta-path $P$.

\end{definition}

\begin{table*}[t]
    \centering
      \caption{The description of each meta-path}
      \begin{tabular}{cccc }
      \toprule
       \textbf{DID} & \textbf{Meta-path} & \textbf{Matrix} & \textbf{Description of each element in Matrix}  \\
       \midrule
       1 & \textit{drug} $\xrightarrow{T}$ \textit{protein} $\xrightarrow{T^{T}}$ \textit{drug} & $TT^T$ & number of common target proteins drug $i$ and $j$ have
        \\
        2 & \textit{drug} $\xrightarrow{T}$ \textit{protein} $\xrightarrow{P}$ \textit{protein} $\xrightarrow{T^{T}}$ \textit{drug} & $TPT^T$ & number of common hierarchical protein interactions drug $i$ and $j$ have
        \\         
        2 & \textit{drug} $\xrightarrow{H}$ \textit{chemical sub-structure} $\xrightarrow{H^{T}}$ \textit{drug} & $HH^T$ & number of common chemical substructures shared by drugs $i$ and $j$
        \\       
        3 & \textit{drug} $\xrightarrow{C}$ \textit{side-effect} $\xrightarrow{C^{T}}$ \textit{drug} & $CC^T$ & number of common side-effects caused by drugs $i$ and $j$
        \\       
         
        \bottomrule
        \end{tabular}
        \label{tab2}
     \end{table*}

In our proposed model, we consider four different meta-paths, as listed in Table ~\ref{tab2}, between drugs:

\begin{enumerate}
\item DID-1: Drug - Protein - Drug 
\item DID-2: Drug - Protein - Protein - Drug 
\item DID-3: Drug - Chemical Sub-structure - Drug 
\item DID-4: Drug - Side Effect - Drug
\end{enumerate} 

We can enumerate a large number of meta-paths given a system architecture with various sorts of entities and their interactions. In our application, we design four valid meta-paths as listed in Table ~\ref{tab2} for similarity measures over drugs based on the obtained data and the four distinct types of relationships mentioned in Table ~\ref{tab1}. 
Different meta-paths assess the similarities between two drugs from various perspectives. The DID-1 meta-path calculates the similarity of two drugs based on their common target proteins. If two drugs have same target protein, there will be path between them coming though that protein. Further, DID-2 meta-path calculates the similarity of two drugs based on their hierarchically common target proteins. For this meta-path, we take both drug-protein and protein-protein interactions into account. If there is a protein that interacts with other drug's target protein, then there will be path between drugs connected through their target proteins. Moreover, DID-3 connect two drugs based on their shared chemical substructures, which indicates how structurally similar the two drugs are. Chemical substructures of drugs are represented as SMILES strings.  A string of a drug is converted into MACCS keys, a binary fingerprint consisting of 167 keys. Every bit position corresponds to a unique chemical substructure, indicating whether or not it is present. We consider each unique chemical substructure as a unique node in the graph. If two drugs share same chemical substructure, there will be path between them coming though that substructure. 
Last but not least, DID-4 weighs the relatedness of two drugs based on the common side effects they cause. If two drugs case same side-effect, there will be path between them coming though that side-effect. As a result, HIN can naturally provide us different similarities with different meta-path-based semantics.

A meta-path between two drugs can be created using relation matrix matrix defined above. 
For example, the DID-1 meta-path between two drugs can be created as \textit{drug} $\xrightarrow{target}$ \textit{protein} $\xrightarrow{target^{T}}$ \textit{drug}.
The commuting matrix for this meta-path is computed by $ T $ * $ P $ *$ T^T $, where $ P $ is the adjacency matrix indicating protein-protein interactions, $ T $ is the adjacency matrix between drugs and target proteins, and $ T^T $ is the transpose of $ T $.

\begin{table}[t]
    \centering
      \caption{Notations and Explanations of \texttt{HAN-DDI}}
      \begin{tabular}{cc}        
      \toprule
       \textbf{Explanation} & \textbf{Notation} \\
       \midrule
        Meta-path  & $\nu$
        \\         
        Initial node feature & $h$
        \\         
        Type-specific transformation matrix & $M_\nu$
        \\       
        Projected node feature & $h^{'}$
        \\       
        Importance of meta-path based node pair $(i,j)$ & $e^{\nu}_{i,j}$
        \\       
        Weight of meta-path based node pair $(i,j)$  & $\alpha^{\nu}_{i,j}$
        \\       
        Meta-path based neighbors & $N^\nu$
        \\ 
        Semantic-specific node embedding & $z^\nu$
        \\       
        Semantic-level attention vector & $q$
        \\       
        Importance of meta-path $\nu$ & $w^{\nu}$
        \\       
        Weight of meta-path $\nu$  & $\beta^{\nu}$
        \\       
        The final embedding & $Z$
        \\
        \bottomrule 
        \end{tabular}
        \label{tab5}
     \end{table}
     
Now, to cope with arbitrary graph-structured data, graph neural network models has been developed. However, they are all intended for homogeneous networks. Because meta-paths and meta-path-based neighbors are two key features in a heterogeneous graph, we will now describe \texttt{HAN-DDI}, intended for heterogeneous graph data that may utilize the subtle differences between nodes and meta-paths. Table ~\ref{tab5} summarizes the notations we will use throughout the article.

\subsection{Feature Extraction}
One of its biggest strengths of graph neural networks (GNNs) is including node features into the learning process. In a general GNN architecture, initialized node features are used to generate enriched and effective node embeddings through a message passing mechanism. Therefore, we need featured of nodes as drugs, proteins and side effects to give to the GNN model in the next step. Here, we describe how we construct features for drugs.

We use the chemical substructures of drugs to create features of them. They are represented as SMILES strings. We employ the ESPF algorithm to extract significant chemical substructures from drugs’ SMILES strings and use them to create drug features.

ESPF is an effective technique for decomposing sequential structures into interpretable functional groups. A few substructures are responsible for chemical reactions in drugs. Thus the algorithm selects the most common substructures as significant. ESPF decomposes a SMILES string $S$ into a set of different sized frequent substructures, beginning with the set of all atoms and bonds. If the frequency of each substructure, $f_i$ in $S$, exceeds the predetermined threshold, then it is added to a list of substructures as a vocabulary $F = f_1....f_n$. This process is repeated until there are no pairs with a frequency more significant than the threshold or the list size exceeds the maximum size.

\subsection{ Heterogeneous Graph Attention Network} 
We utilize the Heterogeneous graph attention Network (HAN-DDI) for DDI prediction by following the idea HAN~\cite{Wang2019HeterogeneousGA}. HAN-DDI includes an encoder, which generates the embedding of drugs, and a decoder that uses the embedding of drugs from the encoder to predict whether a drug pair interact or not.

\subsubsection{Encoder: Drug representation learning}
The encoder layer uses weighted neighborhood aggregation to build a node embedding. First, we define node-level attention to learn the weight of meta-path-based neighbors and aggregate them to get the meta-path-specific node embedding. Then we define the meta-path level attention to learn the importance of meta paths and their weights to combine multiple meta-path-specific drug embeddings into one drug embedding. 

\textbf{Node Level Attention} Meta-paths are used by our model to propagate information from local neighbors. Now, all meta-path-based neighbors might not affect a target node the same way. Each node’s meta-path-based neighbors play a varied role and so have a different impact on learning the node’s embedding. We incorporate node-level attention for each meta-path to learn node representations and then combine the representations of these significant neighbors to produce a final node embedding.

Due to the fact that nodes are of different types, various types of nodes have diverse feature spaces. As a result, we create the type-specific transformation matrix $M_i$ for each kind of node (e.g., node of type $\nu_i$) to project the features of various types of nodes into the same feature space. The projection procedure is as follows:

\begin{equation} 
\label{eq0}
\begin{split}
h_i^{'} & = M_{\nu_{i}} \circ h_i \\
\end{split}
\end{equation}

Here, $h$ and $h'$ represent the original and projected features of nodes $i$ and $\circ$ denotes inner product among two matrices.

Given a node pair $(i,j)$ connected by a meta-path $\nu$, the node-level attention $e^{\nu}_{ij}$ can determine how significant node $j$ will be for node $i$. The significance of the meta-path-based node pair $(i,j)$ can be expressed as follows:

\begin{equation} 
\label{eq1}
\begin{split}
e^{\nu}_{ij} & = att_{node} (h_i^{'}, h_j^{';} \nu) \\
\end{split}
\end{equation}

$Att_{node}$ signifies the deep neural network that conducts node-level attention. Given a meta-path, $Att_{node}$ is shared by all meta-path-based node pairings. Using masked attention, we infuse structural information into the model where we compute $e^{\nu}_{ij}$ for nodes $j$ $\in$ $N^\nu_i$, where $N^\nu_i$ indicates the meta-path-based neighbors of the node $i$. (include itself). Furthermore, we normalize the weights for all meta-path-based neighbors after collecting the weights to derive the attention coefficient $\alpha_{ij}$ using a Softmax function.

\begin{equation} \label{eq2}
\begin{split}
\alpha^{\nu}_{ij} & = softmax_j(e^{\nu}_{ij}) \\
\end{split}
\end{equation}

Then, in the following $(i+1)$-layer, the meta-path-based embedding of node $i$ may be substantially aggregated by the neighbor's embeddings at $i$-layer with the relevant attention coefficients as follows:

\begin{equation} \label{eq3}
\begin{split}
z^{\nu}_{i} & = \delta( \sum_{j \in N^\nu(i)} (\alpha^{\nu}_{ij} \circ h_j^{'} ) ) \\
\end{split}
\end{equation}

where $z^{\nu}_{i}$ is the meta-path $\nu$ learnt embedding of node $i$ and $\delta$ is a non-linear activation function,i.e., RELU. Multi-head attention is used to make the learning process of self-attention more resilient. Specifically, K attention mechanisms are utilized individually to accomplish the feature transformation stated by Eq. ~\ref{eq3}, and then the modified features are concatenated (symbolized as $||$), resulting in the output feature representation as a vector given below.

\begin{equation} \label{eq4}
\begin{split}
z^{\nu}_{i} & = ||_{k=1}^K \delta( \sum_{j \in N^\nu(i)} (\alpha^{\nu}_{ij}  \circ h_j^{'} ) ) \\
\end{split}
\end{equation}

This enables the model to dynamically apply greater aggregate weights to nearby nodes more relevant to the DDI prediction task. As a result, the embedding of the nodes may be aggregated based on the dynamic weight. Because of these properties, our technique is very effective for representation learning.

textbf{Meta-path Level Attention} In various networks, each node (i.e., drug) may have a variety of meta-path information. We use a meta-path-level attention method to combine several meta-path-specific representations for each node. We learn the weight of each meta-path $\nu$ based on the following equation:

\begin{equation} \label{eq5}
\begin{split}
w^{\nu} & =  \sum_{i\in V} q^T tanh (W \circ z^{\nu}_{i} +b ) \\
\end{split}
\end{equation}

where $W$ is the weight matrix, $b$ denotes the bias vector, and $q$ denotes the meta-path-level attention vector. Finally, we use the softmax function to normalize the attention scores for a  meta-path $\nu$ as in Equation ~\ref{eq6}.

\begin{equation} \label{eq6}
\begin{split}
\beta^{\nu} & = \frac{exp(w^{\nu})}{\sum_{t=1}^{T} w^{t} } \\
\end{split}
\end{equation}
where $T$ is the number of meta-paths. The final representation for each node $i$ is then obtained by aggregating the meta-path-specific representations as follows:

\begin{equation} \label{eq7}
\begin{split}
z_i & = \sum_{t=1}^{T} \beta^{t} z_i^{\nu} \\
\end{split}
\end{equation}

\subsubsection{Decoder: DDI learning}
Our objective is to learn whether drug pairs interact using the representation of drugs obtained from the encoder. Decoder, in particular, assigns a score to drug pair ($v_i$,$v_j$) expressing how probable it is that drug $v_i$ and $v_j$ are interacting. We use the dot predictor function as a decoder:

After performing an element-wise dot product between the corresponding drug's features, we obtain the scalar score for each edge.
\begin{equation}
\gamma(z_x,z_y) = z_x \cdot z_y.
\end{equation}

Following that, we put the decoder output into a sigmoid function. 
$y_{x,y} =\sigma({\gamma(z_x,z_y)})$
that generates a prediction score, $Y'$, ranging from 0 to 1. A score close to 1 indicates that there is a high likelihood of interaction between two drugs, whereas a score close to 0 indicated that interaction is less likely.

\subsubsection{Model Training}
 We train our entire encoder-decoder architecture as a binary classification problem by minimizing a binary cross-entropy loss function specified as
\begin{equation}
L= -\sum_{i=1}^{N}{Y_{i}\log Y_{i}^{'}+(1-Y_{i})\log(1-Y_{i}^{'})}
\end{equation}
where $N$ is the total number of samples, $Y_{i}$ is the actual label and $Y_{i}^{'}$ is the predicted score.

\section{Experimental Results}
\label{experiment}
In this section, we evaluate the performance of our proposed model, \texttt{HAN-DDI} for DDI prediction with extensive experiments and compare the results with the state-of-art baseline models using several accuracy metrics. The model used to predict the DDIs of existing drugs may not be as effective as the model used to predict the DDIs of new drugs. Therefore, we assess our model’s performance for new and existing drugs. First, we describe how we collect our data, then we explain our experiments, and finally, we analyze our results.

\subsection{Data Collection}
We create the multi-type DDI dataset by gathering DDI items from DrugBank. We choose 513 approved drugs and obtain their chemical structures, side effects, and drug binding proteins. We obtain 11,845 interactions among 513 drugs.  Moreover, we extract 413 PPI interactions from BioGRID. Table~\ref{tab3} summarizes the fundamental statistics of the dataset used.
\begin{table}
    \centering
      \caption{Statistics of Dataset}
      \begin{tabular}{cc }
      \toprule
       \textbf{Nodes/Edges} & \textbf{Number of nodes} \\
       \midrule
        Drug  & 513
        \\         
        Protein & 290
        \\         
        Side Effect & 527
        \\       
        DDI & 11845
        \\       
        DPI & 514
        \\       
        Drug-Side Effect & 13674
        \\ 
        PPI & 413
        \\       
        \bottomrule
        \end{tabular}
        \label{tab3}
     \end{table}

\subsection{Parameters Used}
We employ an end-to-end optimization method for \texttt{HAN-DDI}, simultaneously optimizing all trainable parameters and propagating loss function gradients via both encoder and decoder for implementing DDI prediction. To optimize the model, we use the Adam optimizer with a learning rate of 0.005 and a dropout rate of 0.6 for a maximum of 20 epochs (training iterations). Moreover, the number of heads and hidden units used are 8 and 16, respectively. The parameters used in this model are outlined in Table~\ref{tab4}.

\begin{table}[t]
    \centering
      \caption{Hyper-parameter Settings}
      \begin{tabular}{cc}        
      \toprule
        Learning rate  & 0.005
        \\         \hline
        Number of heads & 8
        \\         \hline
        Hidden units & 8
        \\       \hline
        Dropout & 0.6
        \\       \hline
        Weight decay & 0.001
        \\       \hline
        Number of epochs & 200
        \\       \hline
        Patience & 100
        \\       
        \bottomrule 
        \end{tabular}
        \label{tab4}
     \end{table}

\subsection{Baseline methods}
We compare our method with the following state-of-the-art methods. For the model that generate drug node embedding, we apply concatenation and use the concatenated embedding as the feature of drug pairings. Lastly, we feed concatenated embedding to a machine learning classifier. Here we apply different  ML models and select the decision tree classifier to further experiments. Detail of the baselines are summarized bellow based on their types.

\begin{itemize}
\item \textit{Graph Embedding on DDI:} Graph embedding models generates the representation of nodes in a homogeneous graphs based on neighborhood information. We use DeepWalk \cite{Perozzi2014DeepWalkOL} and node2vec \cite{Grover2016node2vecSF} as two main approaches for graph embedding based on a random walk. A low-dimensional feature representation of drug nodes is created these models on the DDI networks, which only include drug drug interaction information. 

\item \textit{Graph Neural Network on DDI:} We use GNN architectures on DDI graphs to learning representation of drugs. We select three common GNN-based methods: GCN\cite{Kipf2017SemiSupervisedCW}, GAT\cite{Velickovic2018GraphAN}, and GraphSAGE \cite{Hamilton2017InductiveRL}. These GNN models are obtained from DGL\footnote{https://docs.dgl.ai/}. 

\item \textit{Graph Neural Network on Heterogeneous Graph:} We use common GNN architectures on our heterogeneous graphs to learning representation of drugs. We use  same 3 GNN models, GCN, GAT and GraphSage, to learn node embedding.

\item \textit{Graph Neural Network on Homogeneous Graphs:} We construct various homogeneous graphs consisting of drug nodes, where edges among drug nodes are constructed  based on their relation to other entities as if they share target proteins, cause side effects, or possess similar chemical substructures. We have 3 different graphs as described below-

\begin{itemize}

\item HG1: Node type: Drugs; Edge: Drug Nodes sharing same the target proteins
\item HG2: Node type: Drugs; Edge: Drug Nodes causing the same side effects
\item HG3: Node type: Drugs; Edge: Drug Nodes possessing the same chemical substructures

\end{itemize}

To learn drug node embedding, we apply GCN to these homogeneous graphs.

\item ML Classifier on Drug Functional Representation (FR):  Principal Component Analysis (PCA)\cite{Shen2009PrincipalCA} is a dimensionality-reduction approach that is commonly used to reduce the dimensionality of huge data sets. This method generates a feature vector for each drug based on the PCA representation of the drug-target protein interaction matrix, the PCA representation of the drug-chemical substructure possession matrix, and the PCA representation of individual drug side effects.

\item Heterogeneous Graph-Based Methods: We use Decagon ~\cite{ZitnikAL18} model for this baseline. This graph convolution network model was created to predict multi-relational links in heterogeneous networks. End-to-end learning is achieved in this model to produce drug embedding using graph convolution and to predict, DDIs using a decoder.

\end{itemize}

\subsection{Comparison with baselines}

\begin{table*}[t]
\centering{
\caption{ \centering{Performance comparisons of HAN-DDI with baseline models for existing drugs } }
\label{tab:3}
\small
\begin{tabular}{ c | c | c c c c }
\toprule
 \textbf{Model}
 & \textbf{Method}
 & \textbf{ F1 }
  & \textbf{ RECALL  }  & \textbf{ PRECISION } & \textbf{ AUROC } 
  \\ 
\midrule
  &{GAT} & 84.72 & 85.07 & 84.36 & 84.65\\
 &\textbf{GCN} & 85.02 & 86.29 & 83.78 & 84.79 \\
{GNN on DDI graph} & {GraphSAGE} & 84.18	& 85.5 & 82.9 & 83.93 \\

\hline
  &\textbf{Node2Vec}  & 79.63 & 78.35 & 80.95 & 88.73 \\
{ GE on DDI graph } &{DeepWalk}  & 78.8 & 78.27 & 79.44 & 88.64\\
\hline

  &{GAT}  & 85.68 & 86.28 & 85.99 & 86.06\\
 &\textbf{GCN} & 88.28 & 89.01 & 85.49 & 89.68\\
{GNN on Heterogeneous graph} &{GraphSAGE} & 87.79 & 88.31 & 86.94 & 85.59\\
\hline

  &{HG1}  & 85.35 & 86.48 & 84.24 & 85.15 \\
 &{HG2} & 84.67 & 87.04 & 82.43 & 84.24\\
{Homogeneous Graphs} &\textbf{HG3} & 86.01 & 84.45 & 87.63 & 86.26\\
\hline

{ ML  Classifier on drugs' FR } &{Concatenated Drug Features}  & 86.19 & 87.58 &  84.55 & 81.23\\
\hline

& Decagon & 89.92 & 88.88 & 90.12 & 92.52\\

Heterogeneous Graph-based methods & \textbf{ \texttt{HAN-DDI} } & \textbf{95.18} & \textbf{96.77} & \textbf{93.65} & 82.17\\
\bottomrule
\end{tabular}}
\end{table*}

\begin{table*}[t]
\centering{
\caption{ \centering{Performance comparisons of HAN-DDI with baseline models for new drugs } }
\label{tab:4}
\small
\begin{tabular}{ c | c | c c c c }
\toprule
 \textbf{Model}
 & \textbf{Method}
 & \textbf{ F1 }
  & \textbf{ RECALL  }  & \textbf{ PRECISION } & \textbf{ AUROC } 
  \\ 
\midrule
  &{GAT} & 72.55 & 74.07 & 73.15 & 72.48\\
 &{GraphSAGE} & 70.88 & 74.61 & 72.93 & 73.18 \\
{ GNN on DDI graph} & \textbf{GCN}	& 72.39 & 75.54 & 73.22 & 74.68\\

\hline
  & DeepWalk  & 68.15 & 68.47 & 70.37 & 68.59 \\
{ GE on DDI graph } &\textbf{Node2Vec} & 68.84 & 67.41 & 69.72 &66.18\\
\hline

  &{GAT}  & 75.33 & 76.42 & 75.29 & 76.92\\
 &{GraphSAGE} & 76.21 & 75.48 & 75.08 & 74.12\\
{GNN on Heterogeneous graph} & \textbf{GCN} & 77.40 & 79.92 & 75.63 & 78.34\\
\hline

 &{HG1}  & 75.16 & 74.23 & 75.67 & 76.08 \\
 &{HG2} & 74.88 & 76.52 & 73.77 & 75.49\\
{Homogeneous Graphs} &\textbf{HG3} & 76.92 & 75.58 & 76.14 & 75.80\\
\hline

{ML Classifier on drugs' FR} &{Concatenated Drug Features}  & 74.29 & 75.86 & 72.94 & 72.49\\
\hline

Heterogeneous Graph-based methods & \textbf{ \texttt{HAN-DDI} } & \textbf{82.87} & \textbf{84.19} & \textbf{83.75} & 71.48\\
\bottomrule
\end{tabular}}
\end{table*}
We perform detailed experiments on our models and several state-of-the-art models for existing and new drugs. To predict new drugs, we partition our dataset so that 20\% of drugs do not exist in the training set and only appear in the testing set. Instead of concealing just 20\% of drug-drug interactions ~\cite{Gottlieb2012INDIAC,  Vilar2014SimilaritybasedMI}, 20\% of the drugs are selected and all of their DDIs are hidden from the training set. This selected 20\% of drugs considered as new drugs.

Experimental results for existing and new drugs are shown in Tables ~\ref{tab:3} and ~\ref{tab:4}. In prediction for existing and new drugs, \texttt{HAN-DDI} comprehensively surpasses other baselines. For existing drugs, while we get scores are 95.18\%, 96.77\%, 93.65\%, and 82.17\%, respectively, Decagon from baseline with highest score get  89.92\%, 88.88\%, 90.12\% and 92.52\% for for F-$1$ score, Recall, Precision, and AUROC, respectively.
Similarly, for new drugs, while we get scores are 82.87\%, 84.19\%, 83.75\%, and 71.48\%, GCN on Heterogeneous graph from baseline with highest score get  77.40\%, 79.92\%, 75.63\% and 78.34\% for for F-$1$ score, Recall, Precision, and AUROC, respectively.

For further analysis on existing drugs, Node2Vec from graph embedding models performs the best from graph embedding models on DDI graph. GCN provides the best results for GNN models on DDI and heterogeneous graphs. Applying GCN on heterogeneous graphs (HG) produces better results than DDI since more nodes and edges information are considered on HG. In the case of homogeneous graphs (HG), HG3 (homogeneous graph where edges are formed in case drug nodes share similar chemical substructures) produces better accuracy results than other HG. It signifies that chemical substructures are vital in representing drug node embedding and characterizing DDIs. In addition, HG1 (homogeneous graph where edges are formed in case drug nodes share same target proteins) generates better results than HG2 which shows that target protein is more important tan side effects for DDI prediction. On the other hand, the best-performing model among all baselines, Decagon, earns an F1 score of 89.92\%. However, in some cases, \texttt{HAN-DDI} outperforms baseline methods by 19\% performance gain.

For new drugs, we could not perform experiments for Decagon since it operates in a transductive manner, where the model is familiar with all of the data. However, Node2Vec from graph embedding models outperforms DeepWalk. While GCN produces the best results for GNN models on DDI graphs, it is worse than GCN on heterogeneous graph because the graph considers more diverse entity and relational information than DDI graph. Moreover, GCN on heterogeneous graph performs the best across the baseline models.  In the case of homogeneous graph, similar to existing drugs, HG3 exceeds other homogeneous graph variants in evaluation metrics. 

As summary, in most circumstances, \texttt{HAN-DDI} stays superior to all baseline methods for existing and new drugs and, in some cases, outperforms baseline approaches by 17\%.
According to the overall performance, \texttt{HAN-DDI} is either the best or performs close to the best model with respect to different evaluation measures. Furthermore, we investigate whether and under what conditions \texttt{HAN-DDI} outperforms all baseline approaches. Meanwhile, Decagon is widely used as a baseline for any DDI prediction task. We will analyze \texttt{HAN-DDI} method with that of Decagon's. First, the reason behind the outstanding performance achieved by DECAGON is its ability to learn drug representations in a transductive manner. Transductive learning algorithms have previously seen all of the data, including the training and testing datasets. However, since \texttt{HAN-DDI} can predict both existing and new drugs, it demonstrates \texttt{HAN-DDI}'s ability to work in both a transductive and inductive manner. Secondly, Decagon is built on a heterogeneous graph containing different types of nodes and edges. In the encoder part, they maintain a weight matrix for nodes only. \texttt{HAN-DDI} incorporates both node-level and meta-path-level attention, resulting in effective drug embedding. To our knowledge, this is the first work on DDI prediction, which considers both node-level and meta-path-level attention. All these aspects enable \texttt{HAN-DDI} to manifest superior performance compared to any baseline methods.

\begin{figure*}
    \centering
    \includegraphics[width=0.9\textwidth]{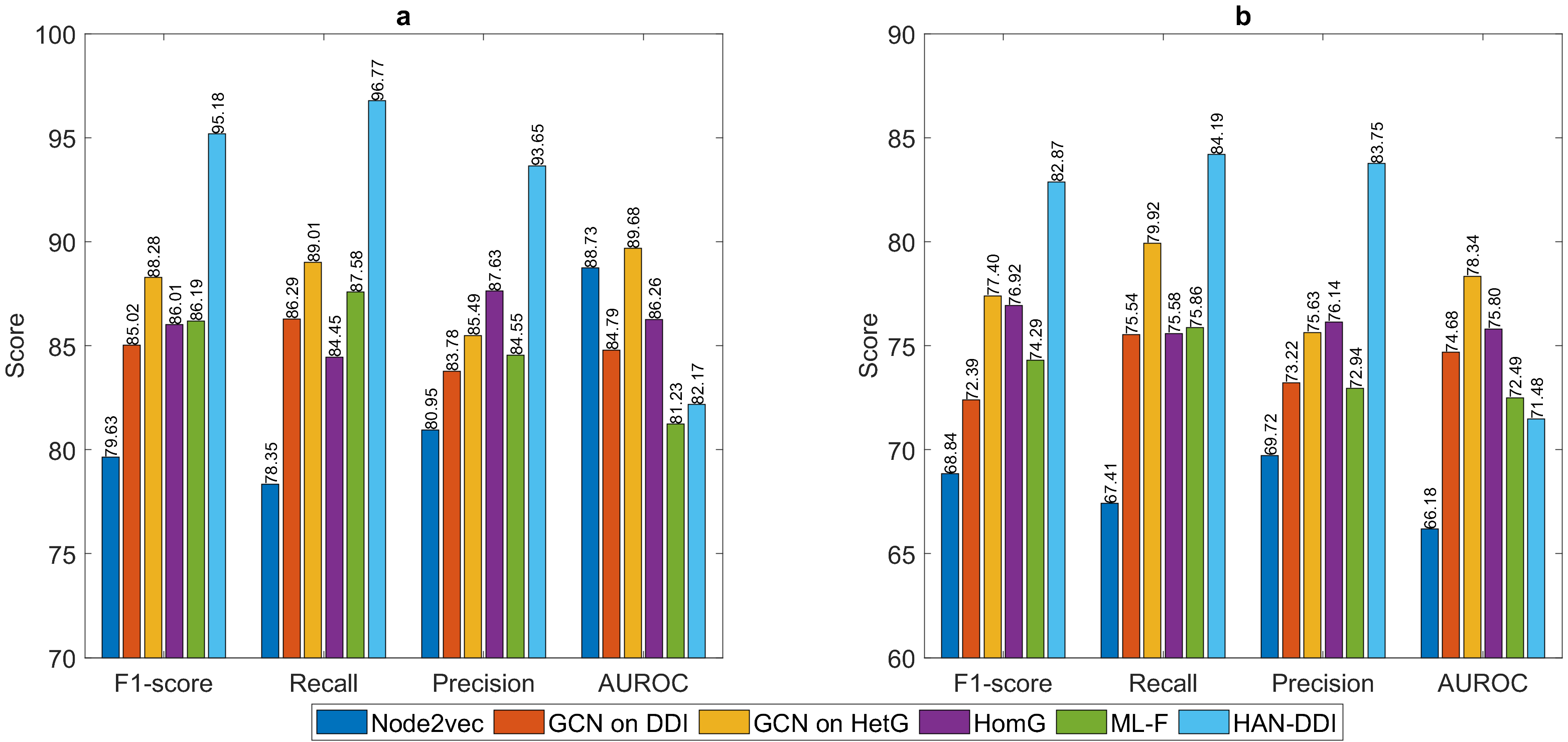}
   \caption{Performance Comparison of Models for a) Existing and b) New Drugs}
    \label{fig:result}
\end{figure*}

Furthermore, to just compare the model with respect to its type,  we choose the best-performing method from each baseline model, which is Node2Vec from graph embedding, HG3 from homogeneous graphs, GCN from GNN on DDI, GCN from GNN on Heterogeneous Graph and Concatenated Drug Features from ML on Functional Representation. Figure ~\ref{fig:result} represent these results.  The figure shows that \texttt{HAN-DDI} is the best-performing model. It is worth noting that, based on our findings, all GNN-based models, including HAN-DDI and baselines, generates magnificent results. Because of its capacity to analyze graph structure data, Graph Neural Network (GNN) has lately gained a lot of interest. It is an effective tool for analyzing graph data. The capability of GNN to represent the interactions between graph nodes is a milestone in graph analysis research. Moreover, message passing between graph nodes allows GNNs to capture graph dependence. Heterogeneous graph Attention network is a variant of GNN. In \texttt{HAN-DDI}, we incorporate a heterogeneous graph attention network-based encoder. It incorporates both node-level and meta-path-level attention, enabling us to learn the weight of each drug and meta-paths. Since this model allows us to generate enriched drug embedding through node-level and meta-path-level attention, it produces superior accuracy results compared to GNN-based models.
This model is very efficient and may be utilized well for different machine learning tasks including node classification and link prediction. 

\subsection{Case Study: Prediction and Validation of Novel DDI Predictions}

\begin{table}
    \centering
      \caption{Novel DDI Predictions and Their Validation}
      \begin{tabular}{cccc }
      \toprule
       Drug1 & Drug2 & Predicted Score & DrugBank Label \\
       \midrule
Quinolones & Macrolides & 0.996 & 1\\
Phenobarbital & Rifampin & 0.972 & 1\\
Sildenafil & Cimetidine & 0.958 & 1\\
Carbamazepine & Cimetidine & 0.932 & 1 \\
Quinolones & Citalopram & 0.91 & 1\\      
        \bottomrule
        \end{tabular}
        \label{Tab:6}
     \end{table}

 Our goal is to assess the accuracy of HAN—DDI regarding DDI prediction with real DDI instance. We compare our DDI predictions with DrugBank labeled data. We generate different drug pairs and then predict a score these drug pairs. For the top rank pairs that are likely to cause DDIs, we search DrugBank to see if our prediction can be found as DDI in Drugbank.

Table~\ref{Tab:6} displays \texttt{HAN-DDI}'s prediction score and the DrugBank’s labeled data for corresponding drug pairs. We exhibit that our top 5 DDI predictions can be found in DrugBank. Therefore, experimental results show that our model can accurately identify drug pairs likely to cause DDI.  For example, \texttt{HAN-DDI} predicts that simultaneous use of Cimatedine and Carbamazepine can cause DDI. This experiment illustrates \texttt{HAN-DDI}'s ability to predict novel DDI predictions.

\begin{figure}[t]
    \centering
    \includegraphics[width=0.45\textwidth]
    {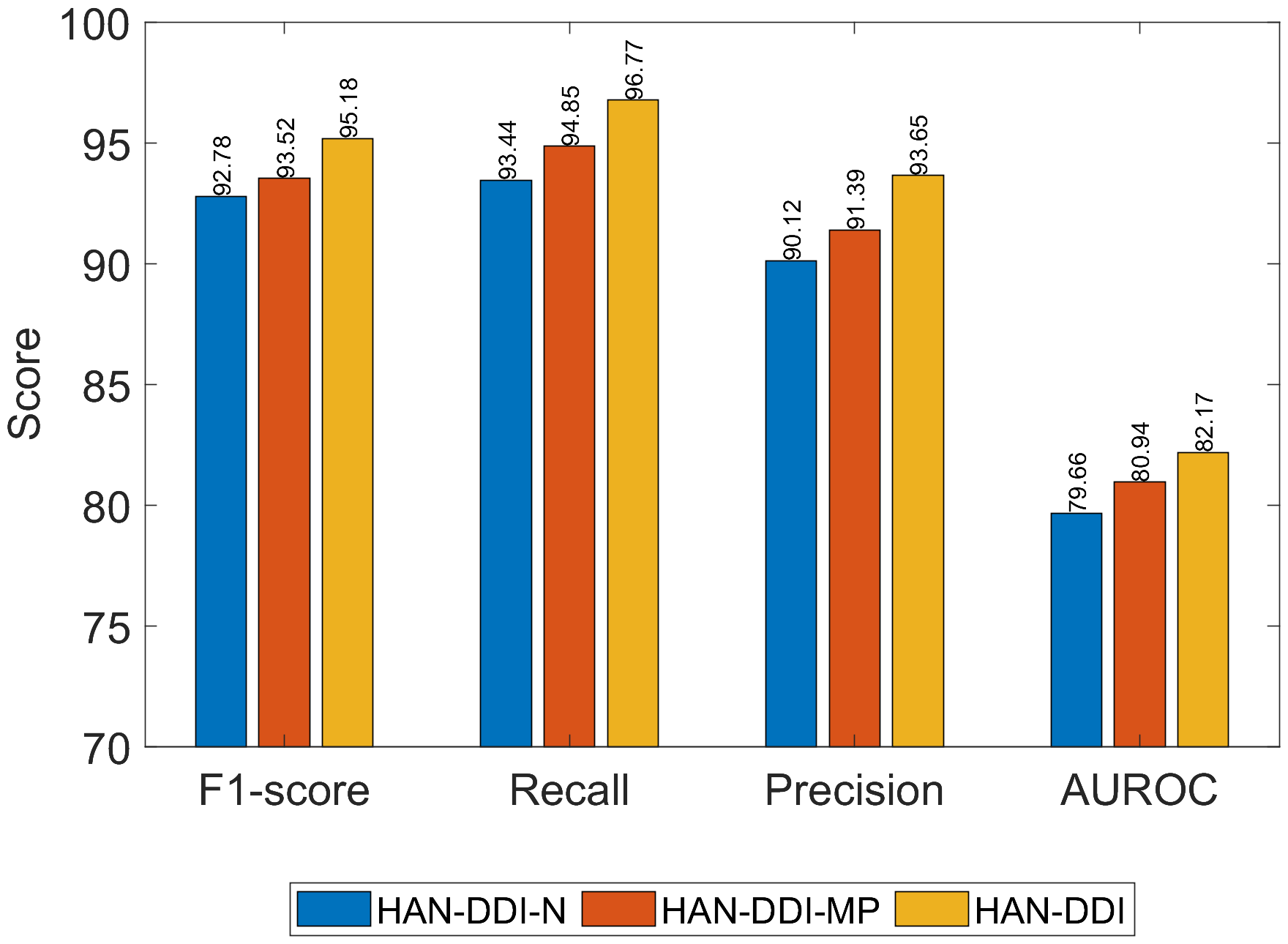}
   \caption{Performance Comparison of HAN-DDI with its variants}
    \label{fig:HAN-DDI variants}
\end{figure}

\subsection{Detailed Analysis On Node-level and Meta-path-level Attention}
Our \texttt{HAN-DDI} model incorporates two types of attention: node-level attention and meta-path-level attention. These two levels of attention aim to understand the attention of meta-path-specific neighbors and meta-path, respectively. In this section, we
run an experiment to assess their influence on overall performance and develop the following \texttt{HAN-DDI} variants:
\begin{itemize}
\item \texttt{HAN-DDI-MP}: We use meta-path-level attention acquired from \texttt{HAN-DDI}. A random matrix is used for node attention rather than the node-level attention matrix.

\item \texttt{HAN-DDI-N}: We use node-level attention generated by \texttt{HAN-DDI}. For meta-path-level attention, we assign equal weight to each meta-paths.
\end{itemize}

For \texttt{HAN-DDI-MP}, the average F$ 1$-score, Recall, Precision, and AUROC scores  are 93.65\%, 94.18\%, 90.72\%, and 80.88\%, respectively, for existing drugs. For \texttt{HAN-DDI-N}, our average F$ 1$-score, Recall, Precision, and AUROC scores are 92.87\%, 93.19\%, 89.75\%, and 78.48\%, respectively.

Performance analysis of \texttt{HAN-DDI-MP} and \texttt{HAN-DDI-N} is depicted in Figure ~\ref{fig:HAN-DDI variants}, which indicates that \texttt{HAN-DDI} outperforms \texttt{HAN-DDI-MP} and \texttt{HAN-DDI-N}. It refers to the fact that node-level and meta-path-level attention are effective in retrieving various meta-path information of nodes in different meta-paths. However, meta-path-level attention is more influential than node-level attention since \texttt{HAN-DDI-N} performs worse than \texttt{HAN-DDI-MP}.

\section{Conclusion}
\label{conclusion}

In this paper, we present \texttt{HAN-DDI}, a novel graph encoder-decoder based on heterogeneous graph neural networks for predicting DDIs. We integrated a two-level attention method, i.e., node-level and meta-path-level attention, that can efficiently determine the importance of meta-path-specific neighbors and meta-paths to exploit various meta-path information of nodes from heterogeneous networks fully. Extensive experimental results show that the suggested \texttt{HAN-DDI} model is dependable and promising in predicting DDIs. Apart from generating outstanding accuracy results, we can show that our model can accurately bring out the drug pairs likely to cause DDI by comparing our prediction with real-world evidence. Our approach can be used to solve more complex Bioinformatics problems such as polypharmacy-side effect prediction.
\section*{Acknowledgment}
This work is funded partially by National Science Foundation under Grant No 2104720

\bibliographystyle{ACM-Reference-Format}
\bibliography{sample-base}

\end{document}